\documentclass[nohyperref]{article}

\usepackage{microtype}
\usepackage{graphicx}
\usepackage{subfigure}
\usepackage{booktabs} 

\usepackage{hyperref}

\usepackage[accepted]{icml2022}

\usepackage{amsmath}
\usepackage{amssymb}
\usepackage{mathtools}
\usepackage{amsthm}
\usepackage[utf8]{inputenc} 
\usepackage[T1]{fontenc}    
\usepackage{booktabs}       
\usepackage{amsfonts}       
\usepackage{nicefrac}       
\usepackage{microtype}      
\usepackage{xcolor}         
\usepackage{algorithm}
\usepackage{amsmath}
\usepackage{amssymb}
\usepackage{graphicx}
\usepackage{wrapfig}
\usepackage{hyperref}       

\hypersetup{colorlinks,linkcolor={blue},citecolor={blue},urlcolor={blue}}
\newcommand{\Weight}{\ensuremath{\mathbf{W}}}
\newcommand{\iWeight}[1]{\ensuremath{\Weight_{#1}}}
\newcommand{\Bias}{\ensuremath{\mathbf{B}}}
\newcommand{\iBias}[1]{\ensuremath{\Bias_{#1}}}

\newcommand{\iact}[2]{\ensuremath{\varphi_{#2}(#1)}}
\newcommand{\Error}{\ensuremath{\mathbf{\Delta}}}
\newcommand{\iError}[1]{\ensuremath{\Error_{#1}}}
\newcommand{\bbR}[1]{\ensuremath{\Re^{#1}}}
\newcommand{\IntAct}{\ensuremath{\mathbf{Z}}}
\newcommand{\iIntAct}[1]{\ensuremath{\IntAct_{#1}}}
\newcommand{\OutAct}{\ensuremath{\mathbf{A}}}
\newcommand{\iOutAct}[1]{\ensuremath{\OutAct_{#1}}}

\newcommand{\grad}[2]{\ensuremath{\nabla_{#2}#1}}
\newcommand{\Loss}{\ensuremath{\mathcal{L}}}
\newcommand{\bfvec}[1]{\mathbf{#1}}
\renewcommand{\Re}{\mathbb{R}}

\usepackage[capitalize,noabbrev]{cleveref}

\theoremstyle{plain}

\theoremstyle{definition}

\theoremstyle{remark}

\usepackage[textsize=tiny]{todonotes}

\icmltitlerunning{Peering Beyond the Gradient Veil with Distributed Auto Differentiation}

\begin{document}

\twocolumn[
\icmltitle{Peering Beyond the Gradient Veil with Distributed Auto Differentiation}



\icmlsetsymbol{equal}{*}

\begin{icmlauthorlist}
\icmlauthor{Bradley T. Baker}{trends,gsu,gatech}
\icmlauthor{Aashis Khanal}{trends,gsu}
\icmlauthor{Barak A. Pearlmutter}{mayn}
\icmlauthor{Vince D. Calhoun}{trends,gsu,gatech}
\icmlauthor{Sergey M. Plis}{trends,gsu}
\end{icmlauthorlist}
 
\icmlaffiliation{trends}{Tri-Institutional Research Center for Neuroimaging and Data Science (TReNDS)}
\icmlaffiliation{gsu}{Georgia State University}
\icmlaffiliation{gatech}{Georgia Institue of Technology}
\icmlaffiliation{mayn}{Maynooth University}

\icmlcorrespondingauthor{Bradley Baker}{bbaker43@gsu.edu}

\icmlkeywords{Machine Learning, ICML}

\vskip 0.3in
]



\printAffiliationsAndNotice{}  

\begin{abstract}
Although distributed machine learning has opened up many new and exciting research frontiers, fragmentation of models and data across different machines, nodes, and sites still results in considerable communication overhead, impeding reliable training in real-world contexts.
   The focus on gradients as the primary shared statistic during training has spawned a number of intuitive algorithms for distributed deep learning; however, gradient-centric training of large deep neural networks (DNNs) tends to be communication-heavy, often requiring additional adaptations such as sparsity constraints, compression, and quantization, to curtail bandwidth.
   We introduce an innovative, communication-friendly approach for training distributed DNNs, which capitalizes on the outer-product structure of the gradient as revealed by the mechanics of auto-differentiation. The exposed structure of the gradient evokes a new class of distributed learning algorithm, which is naturally more communication-efficient than full gradient sharing. Our approach, called distributed auto-differentiation (dAD), builds off a marriage of rank-based compression and the innate structure of the gradient as an outer-product. We demonstrate that dAD trains more efficiently than other state of the art distributed methods on modern architectures, such as transformers, when applied to large-scale text and imaging datasets. The future of distributed learning, we determine, need not be dominated by gradient-centric algorithms. 

\end{abstract}

\section{Introduction}
\label{sec:intro}

The distributed deep learning community has long gravitated towards methods which share gradients during training \cite{bottou2010large, verbraeken2020survey}. Owing in part to the linearity of the gradient, methods like distributed stochastic gradient descent (dSGD) have served as the backbone for large-scale frameworks such as horovod~\cite{sergeev2018horovod}, PyTorch~\cite{paszke2019pytorch}, and others. When viewed through the lens of auto-differentiation (AD) \cite{Speelpenning1980}, however, we can easily observe that the gradient is computed as the outer-product of two smaller matrices, which are accumulated during the forward and backward passes through the network. In this work, we develop this simple fact into an elegant new framework for distributed deep learning. We show that methods grounded in AD naturally provide a bandwidth reduction over standard dSGD and other state of the art methods like PowerSGD, along with competitive performance. We aim to show that much can be gained by turning the focus of distributed learning away from gradient-centrism and toward auto-differentiation.

The many  parameters at work in deep neural networks (DNNs) require significant amounts of data to train, with over-fitting becoming a real possibility if not enough data are provided, or the network is not otherwise regularized. The need for training on large amounts of data in reasonable time has led the deep learning community to focus on data-parallel training, where models on different (GPU) processors are synchronously trained on their respective subsets of data~\citep{shallue19_measure_data_parallel} maintaining the same gradient.
Distributed deep learning can also be motivated by a desire to keep local training samples hidden.
For example, the application of deep learning to medical problems which utilize highly personal data such as medical imaging scans or DNA sequences can require models to be trained on samples which cannot be transferred from one data gathering site to another due to legal or ethical considerations. These issues motivate privacy-sensitive toolboxes for distributed learning~\citep{plis2016coinstac}.

One of the main obstacles to the scalability of distributed deep learning is the bottleneck introduced by the large amount of information transmitted over the network during training. Zhang et al. \cite{zhang2020network} showed that in $<$ 100 Gb/s networks, training runtime is significantly worse without compression, preventing linear scaling with workers or model size. Svyatkovskiy et al.~\cite{svyatkovskiy2017training} also showed that runtime increased with network size when training distributed RNNs. In extreme cases, \cite{li2018network} where the number of parameters is much larger than network bandwidth, communication time dominates training time. Finally, efficient communication of statistics during training has been a preoccupation of algorithm designers since the advent of the field (see \cite{dean2012large}\&\cite{sergeev2018horovod} for specific examples, and \cite{tang2020communication} for survey; also methods mentioned in related works). Even if network architectures are ideally constructed, communication will always limit the overall runtime of distributed algorithms.

\begin{figure*}[t!]
  \centering
  \includegraphics[trim={10 14 12 12}, width=\textwidth]{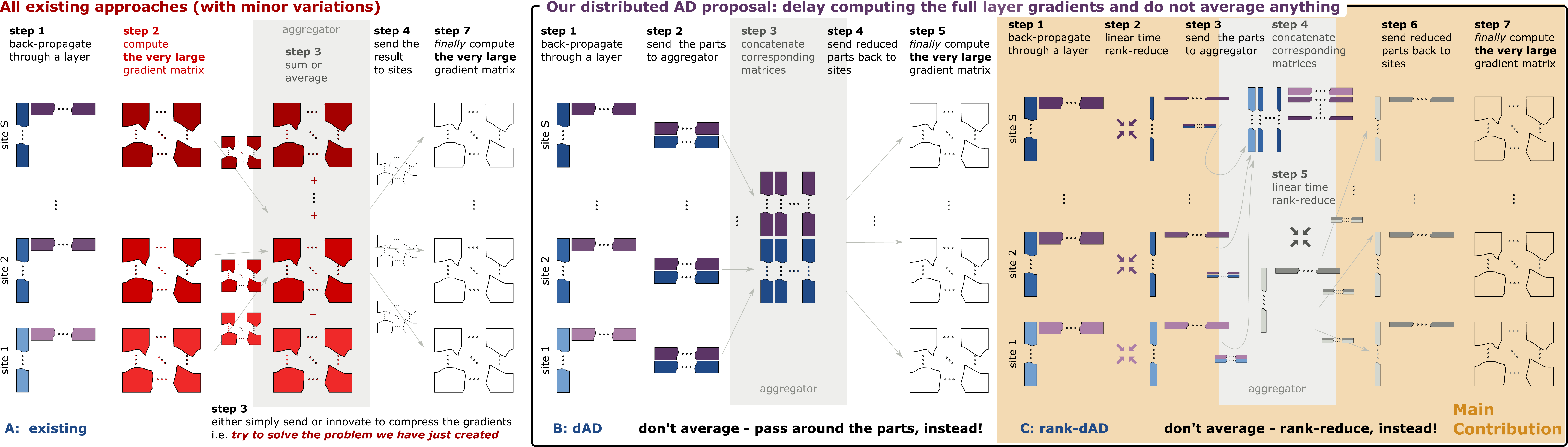}
  \vspace{-10pt}
  \caption{A high level depiction of the state of the art approaches
    to distributed SGD (A), demonstration of our observation that
    working with AD process directly provides bandwidth reduction (B),
    and a linear reduction algorithm that significantly reduces the
    bandwidth (C).}
  \label{fig:diagram}
  \vspace{-15pt}
\end{figure*}

If we take a step back from gradient-centric distributed methods, however, we notice a simple but startlingly profound observation regarding the gradient's structure. In reverse-mode auto-differentiation, the gradient of a layer is computed as an outer-product of the input activations to that layer and the partial derivative with respect to that layer's output. These two component matrices are themselves often smaller in size than the full gradient, and perhaps more importantly, they represent an explicit structure at work behind the gradient computation which deserves consideration on its own ground. In this work, we will aim to show that distributed auto-differentiation algorithms exploiting this inherent outer-product structure exhibit myriad benefits, such as significant bandwidth reduction and performance improvements, when compared to standard dSGD algorithms. 

Our contributions in this work can be summarized as follows: First, we highlight a key observation that the outer product structure of the gradient inspires a class of inherently communication efficient distributed learning algorithms. Next, we present a method which elegantly and naturally arises from the gradient's outer-product structure: rank-distributed Auto Differentiation (rank-dAD). Rank-dAD exploits the gradient's outer-product structure so that rank-reduced estimates of the component matrices can be efficiently communicated to reconstruct an estimate for the true global gradient. Through the use of a new, linear-time algorithm for structured power iterations (SPI), rank-dAD drastically reduces bandwidth while still maintaining model performance. The differences between the existing approaches and our proposed step away from the gradient compression are highlighted in Figure~\ref{fig:diagram}. We illustrate our methods on benchmark deep learning problems such as digit recognition with MNIST, continuous time-series classification with a GRU-RNN on the UEA datasets~\cite{bagnall2018uea}, and large-scale image recognition with a vision-transformer~\cite{dosovitskiy2020image} on the cifar-10 dataset.

\section{Related Work}
\label{sec:related_work}

Most relevant work in distributed deep learning focuses on gradients as the primary shared statistic, and applies techniques such as dropping unnecessary values via sparsification \citep{aji2017sparse, wang2018atomo, stich2018local, wen2017terngrad, sattler2019sparse, sattler2019sparsetern, shi2019distributed}, mapping values into bins via quantization \citep{alistarh2016qsgd, agarwal2018cpsgd, bernstein2018signsgd, horvath2019variance, yu2018gradiveq}, or otherwise compressing gradients  \citep{lin2017compression, koloskova2019compression, mishchenko2019distributed}. Although these methods focus on gradients, the majority of them are applicable to any shared statistic which can be used for learning. Since we focus on the more fundamental question of what statistics are being shared, these methods are potentially synergistic with ours.

A second class of distributed deep learning method reduces bandwidth by following an update schedule, so gradients are not shared for every batch or epoch. These methods, such as bursty aggregation \citep{zhang2017poseidon}, lazy aggregation \citep{chen2018lag}, periodic averaging \citep{haddadpour2019local}, and other scheduling strategies \citep{stich2018local,yu2019parallel, assran2019stochastic}, are again agnostic to the actual statistic shared---as long as the statistic can be used for learning. With the exception of methods which average statistics during training, the methods proposed here should be entirely compatible with any update schedule desired, since the auto-differentiation statistics we share can be used to reconstruct gradients at any point during training.

The method most closely related to that proposed here is PowerSGD \citep{vogels2019powersgd}, which uses a QR decomposition of local gradients to estimate two low-rank matrices which can be used to reconstruct a low-rank approximation of the gradient. Although PowerSGD is a gradient compression method on its face, the sharing of low-rank Q and R matrices does represent a shift away from sharing raw gradients. Indeed, for a chosen low rank~$r$, PowerSGD is able to achieve a bandwidth per layer of $\Theta(r(h_i + h_{i+1}))$ for hidden layer sizes $h_i$ and $h_{i+1}$. PowerSGD, which achieves a bandwidth reduction via the path of QR decomposition rather than auto-differentiation, thus represents a conceptually closest alternative to our method. It will be our goal in this work to illustrate the benefits we receive by taking the path of auto-differentiation both in terms of mathematical intuition, model performance, and bandwidth reduction.

\section{Methods}
\label{sec:methods}

Let $\mathcal{N}(W)$ be a deep neural network with $L$ hidden layers. Let $\mathbf{X} \in \Re^{N \times M}$ be the input batch of data with $N$ samples and $M$ dimensions. Let $\mathbf{Y} \in \Re^{N \times C}$ be the target variable of dimension with $N$ samples and $C$ dimensions. let $h_i$ be the size of the $i$th hidden layer.  For a feed-forward network, the weight matrices are thus $\mathbf{W}_i \in \Re^{h_{i} \times h_{i+1}}$, with $h_0 = M$ and $h_{L+1} = C$. Let $\iact{x}{i}$ be the activation used at layer $i$ in the network, and let $\Loss$ be a given cost-function.

For a feed-forward network, the activations at layer $i$ are thus computed as
\begin{align}
  \iIntAct{i} &= \iOutAct{i-1}\iWeight{i} + \iBias{i}
  &
  \iOutAct{i} &= \iact{\iIntAct{i}}{i}
\end{align}

\subsection{Reverse-Mode Auto-Differentiation}

AD is a class of methods through which derivatives of functions may be calculated during the execution of a code which evaluates that function. The backpropagation algorithm for training deep neural networks is specific case of \textit{reverse-mode} AD, in which derivatives are propagated \textit{backwards} along the data flow graph. This is a two stage process: first a \textit{forward pass} through the function is computed, during which intermediate outputs from expressions within the function are saved and relationships between variables are recorded. After the forward pass, a \emph{backward pass} evaluates the contribution of each intermediate variable to the derivative of the output, retracing and combining intermediate variables and expressions until returning to the input. \citep[For a survey, see][]{baydin2018automatic}.

Following reverse mode AD, we use the chain rule to compute the derivative at each layer. At the output layer, we first compute the gradient of the loss w.r.t.\ the output activations $\grad{\Loss}{\iOutAct{L}}$, and take the Hadamard product with the derivative at the output activation $\varphi_L^{\prime}(\iIntAct{L})$ to compute the gradient w.r.t.\ the output, i.e., $\iError{L}$
\begin{align}
\mathbf{\Delta}_L = \grad{\Loss}{\iOutAct{L}} \odot \varphi_L^{\prime}(\iIntAct{L})
\end{align}

At higher layers (where $i<L$), we can continue to compute these errors as
\begin{align}
\begin{split}
\iError{i} &= (\iError{i+1}\iWeight{i} \odot \varphi_{i}^{\prime}(\iIntAct{i}))\label{eq:local_delta}
\end{split}
\end{align}

At layer $i$, the gradient of the weights $\grad{\Loss}{W_i}$ at layer $i$ can thus be computed exactly as
\begin{align}
\begin{split}
\grad{\Loss}{\iWeight{i}} &= \iOutAct{i-1}^{\top}\iError{i}\label{eq:local_grad}
\end{split}
\end{align}

The key insight offered to us by the reverse-mode AD perspective into deep learning is that in many cases the dimensionality of the intermediate variables accumulated during the forward and backwards passes of AD will be less than that of the gradient. In other words. the gradient is a low rank matrix in most practically relevant cases, and this rank is limited from above by the batch size. For example, consider a matrix-vector product $y=Wx$. If Alice is transmitting the gradient $\grad{\Loss}{W}$ to Bob, where only Alice knows~$x$, we can note that $\grad{\Loss}{W}$ is the outer product of $x$ and $\grad{\Loss}{y}$. If $W \in \bbR{m \times n}$, then these two vectors together have dimensionality $m+n$, and we are often in the regime where $m+n \ll m \times n$.

This observation suggests a novel algorithm for backpropagation via distributed auto-differentiation (dAD). We present this algorithm and some more communication-efficient variants in the following sections.

\subsection{Exploiting Outer-Product Structure for Distributed Learning}

The core of our work relies on a rather simple observation: standard auto-differentiation computes the gradient via an outer-product of two smaller matrices (see equation \ref{eq:local_grad}). This elementary fact encourages a new way of looking at distributed learning in which the gradient's component matrices are shared instead of the full gradient itself, as is standard practice. Indeed, in cases where the batch size is significantly less than the hidden dimension ( i.e. $N << h_i$ ) if we were to simply transfer the two component matrices of the gradient, we would already significantly reduce communication overhead. After communicating component matrices, sites can compute \textit{exact} gradients by merely concatenating in the batch dimension, and computing the product as normal. 

Although the na\"ive circulation of component matrices seems an attractive approach on its own, certain difficulties emerge motivating further work to create a more practically useful distributed learning algorithm. One problem, for example, with fully communicating component matrices is that bandwidth usage is directly tied to the chosen batch size, which can be undesirable in applications where large batch-size is desired, or where we accumulate results over an additional sequence dimension, such as when utilizing recurrent architectures. Our fully-realized method achieves efficient communication by capitalizing on the outer-product structure to perform a structured rank-reduction of the component matrices in linear time. In the next sections, we present our complete algorithm for rank distributed auto-differentiation (rank-dAD), the first in our newly-revealed class of efficient algorithms for distributed auto-differentiation.

\begin{algorithm}[t!]
    \caption{rank distributed auto-differentiation (rank-dAD)}
    \label{euclid-dAD}
    \begin{algorithmic}[1] 
    \INPUT{$\{\mathcal{N}_s\}_s^S, \{\mathbf{X}_s\}_s^S, \{\mathbf{Y}_s\}_s^S$}
        \FOR{site $s$}
            \STATE $\{\iOutAct{i}^{(s)}\}_{i=0}^{L} =$ forward($\mathcal{N}_s$, $\mathbf{X}_s$)
        \ENDFOR{}
        \FOR{hidden layer $i = L$, $0 < i \le L$}
            \FOR{site $s$}
                \IF{i == L}
                    \STATE $\iError{L}^{(s)} = \grad{\Loss}{\iOutAct{L}} \odot \varphi^{\prime}(\iIntAct{L}^{(s)})$
                \ELSE
                    \STATE $\iError{i}^{(s)} = \iError{i+1}^{(s)}\iWeight{i}^{(s)} \odot \varphi_{i}^{\prime}(\iIntAct{i}^{(s)})$
                \ENDIF
                \STATE $\mathbf{Q}_i^{(s)}, \mathbf{G}_i^{(s)} =$ SPI($\iError{i}^{(s)},\iOutAct{i-1}^{(s)}$)
            \ENDFOR{}
            \STATE $\mathbf{Q}_i = $ vertcat($\{\mathbf{Q}_i^{(s)}\}_s^S$)             \hfill\COMMENT{At Aggregator}
            \STATE $\mathbf{G}_i = $ vertcat($\{\mathbf{G}_i^{(s)}\}_s^S$)             \hfill\COMMENT{At Aggregator}
            \STATE $\hat{\mathbf{Q}}_i,\hat{\mathbf{G}}_i =
            $SPI$(\mathbf{Q}_i,\mathbf{G}_i)$ \hfill\COMMENT{see Algorithm~\ref{alg:SPI}}
            \STATE broadcastToSites($\hat{\mathbf{Q}}_i,\hat{\mathbf{G}}_i$)
            \FOR{site $s$}
                \STATE $\grad{}{\iWeight{i}}^{(s)} = \hat{\mathbf{P}}_i^\top\hat{\mathbf{Q}}_i$
            \ENDFOR{}
        \ENDFOR{}
    \end{algorithmic}
  \end{algorithm}
  
\subsection{Rank distributed Auto Differentiation}

The outer-product structure of the gradient invites myriad improvements for distributed learning. In this section, we elaborate on how iterative rank-reduction methods can capitalize off of the outer-product structure, fostering an algorithm in which rank-reduced component matrices can be computed efficiently and communicated instead of the full gradient. These component matrices provide a method for estimating the true global gradient while reducing the overall bandwidth usage from quadratic to linear with respect to the layer size. We call this method rank distributed auto-differentiation (rank-dAD), as it combines iterative rank-reduction with our core insight of distributed auto-differentiation into a single, efficient algorithm.

First, we assume that all sites coordinate the initialization of local copies of the chosen architecture - for example, they can share a choice of random seed and probability distribution when generating initial weights. Each site will maintain a local copy of the model weights in memory, and these models will ultimately have equal weights. For simplicity, each site also shares a set of hyper-parameters, such as learning rate, momentum, number of epochs, etc. In principal, hyper-parameters and even certain architectural elements could be allowed to vary between sites based on local needs; however, such circumstances will not affect the overall performance of the model in terms of its communication and computational benefits compared to standard distributed algorithms, and so we leave these as future work.

For a given batch, rank-dAD first performs the typical forward and backward passes from reverse-mode auto-differentiation, accumulating local activations and partial derivatives as would normally be used for local gradient computation. Next, these component matrices are rank-reduced along the batch dimension. Formally, we begin with
$\iOutAct{i-1} \in \bbR{N \times h_{i-1}}$ and $\iError{i} \in \bbR{N \times h_i}$, and reduce these to matrices $\mathbf{G} \in \bbR{k \times h_{i-1}}$ and $\mathbf{Q} \in \bbR{r \times h_i}$ where $r$ is a chosen natural number designating our maximum target rank. Following rank reduction, we transmit the rank-reduced matrices to a single aggregator, or set of aggregators which will perform the next further reduction step. At aggregator nodes, we then concatenate the received $\mathbf{Q}$ and $\mathbf{G}$ matrices, and perform a final rank-reduction to obtain $\hat{\mathbf{Q}}$ and $\hat{\mathbf{G}}$. Once the reduction has obtained data from all sites in the network, we broadcast the final reduced component matrices, and can locally estimate the gradient from these components. For simplicity, we will only designate one node in the network as the aggregator, as this models the architecture supported by COINSTAC \cite{plis2016coinstac}, our target platform; however, in principal, multiple aggregation stages could be used, as in ring-reduce, or hierarchical communication frameworks for example. We leave the detailed investigation of these alternate communication setups as future work.

It remains for us to show exactly how we can compute the rank-reduced component matrices in an efficient manner. It turns out that the structure of the gradient as revealed by AD provides us with a unique approach to rank-reduction which is computationally efficient, and beneficial for integration in our full rank-dAD framework. In the next section, we will present the details of this algorithm, which we call Structured Power Iterations (SPI).

\begin{algorithm}[t!]
    \caption{\label{alg:SPI} Structured Power Iterations (SPI)}
    \label{euclid-dAD}
    \begin{algorithmic}[1] 
    \INPUT{$\mathbf{\Delta}_i$ $\in$ $\mathbb{R}^{N \times h_{i}}$, $\mathbf{A}_{i-1}$ $\in$ $\mathbb{R}^{N \times h_{i-1}}$, $r$ $\in$ $\mathbb{N}_+$, $n, \theta = 10^{-3}$}
    \STATE $\mathbf{Q} = [], \mathbf{G} = []$
    \STATE $\mathbf{C} = \mathbf{A}_{i-1}\mathbf{A}_{i-1}^T$
    \STATE $\mathbf{B} = \mathbf{\Delta}_i^\top \mathbf{C}$
    \FOR{$j=0, j \le r$}
    \STATE $\mathbf{g}_{0}^{j} \sim \mathcal{N}(0, 1)$
    \FOR{$k=1, k < n$}
    \STATE $\mathbf{g}_{k+1}^{j} = \mathbf{B}\Delta_i \mathbf{g}_k^{j} - \mathbf{Q}(\mathbf{G}^\top \mathbf{g}_k^j) $
    \ENDFOR{}
    \STATE $\mathbf{v} = \mathbf{\Delta}_i \mathbf{g}_k^j$
    \STATE $\sigma^j = \sqrt{\mathbf{v}^\top \mathbf{C} \mathbf{v}}$
    \hfill\COMMENT{May be avoided\textsuperscript{\ref{foot:avoidsigma}}}
    \STATE $\mathbf{q}^j = \mathbf{A}_{i-1}^\top \mathbf{v}/\sigma^j$
    \IF{$||\mathbf{g}^{j-1} - \mathbf{g}^{j}||_2/||\mathbf{g}^{j-1}||_2 \le \theta$}
    \STATE break
    \ENDIF
    \STATE $\mathbf{Q} = $concat($\mathbf{Q}, \mathbf{q}^j$)
    \STATE $\mathbf{G} =$concat($\mathbf{G}, \mathbf{g}^j)$
    \ENDFOR{}
    \STATE return$(\mathbf{Q}, \mathbf{G})$
    \end{algorithmic}
\end{algorithm}

\subsubsection{Structured Power Iterations} \label{sec:spi}
Observe that we can compute the singular vector corresponding to the
dominant singular value of $\grad{\Loss}{W_i}$ by iterating the following
recurrence:
\begin{align}
  \bfvec{g}_{k+1}^1 &= (\grad{\Loss}{\iWeight{i}})^{\top}(\grad{\Loss}{\iWeight{i}})
                    \bfvec{g}_k^1 \label{eq:nabla_iter}
\end{align}
Relying on~\eqref{eq:local_grad}, and pre-computing $\mathbf{C} = \iOutAct{i-1}\iOutAct{i-1}^{\top}$,
 $\mathbf{B} = \iError{i}^{\top}\mathbf{C}$, we can
 instead iterate:
\begin{align}
  \bfvec{g}_{k+1}^1 &= \mathbf{B} \left(\iError{i}\bfvec{g}_k^1\right)\label{eq:reduced_iter}
\end{align}
Compared to the $O(h^2)$ complexity of the iteration~\eqref{eq:nabla_iter}, the complexity of the structured power iteration is just $O(h\times N)$ and since $N\ll h$ for all practical models, it is linear in $h$.
The corresponding singular value $\sigma^1 = \sqrt{\bfvec{v}^T \mathbf{C}
\bfvec{v}}$ (where $\bfvec{v} = \iError{i}\bfvec{g}$) is computed in
$O(h\times N)$ once per singular vector, while computation of the left
singular vector $\bfvec{q}^1$ is just $O(h\times N)$ as $\bfvec{q}^1 =
\iOutAct{i-1}^{\top} \bfvec{v}/\sigma^1$.

We successively collect $(\sigma^j\bfvec{g}^j, \bfvec{q}^j)$,
absorbing singular values into one of the vectors\footnote{\label{foot:avoidsigma}In practice
we bypass computing $\sigma^j$ and gain additional speed up because it cancels out in the outer
product since $\bfvec{q}^j$ contains $1/\sigma^j$ factor. }
constructing respective $\mathbf{G}_j$ and $\mathbf{Q}_j$ by concatenating
$\bfvec{g}$s and $\bfvec{q}$s as columns, and proceed to computing the
next singular vectors set by peeling the previously computed
least-squares optimal low-rank representation:
\begin{equation}
  \begin{aligned}
    \bfvec{g}_{k+1}^j &=
    \left((\grad{\Loss}{\iWeight{i}})^{\top}(\grad{\Loss}{\iWeight{i}})
    - \mathbf{Q}_{j-1}\mathbf{G}_{j-1}^{\top}\right)
    \bfvec{g}_k^j \\
    &=
    (\grad{\Loss}{\iWeight{i}})^{\top}(\grad{\Loss}{\iWeight{i}})\bfvec{g}_k^1
    - \mathbf{Q}_{j-1}(\mathbf{G}_{j-1}^{\top}
    \bfvec{g}_k^j)\label{eq:peel}
  \end{aligned}
\end{equation}
We have already shown that complexity of the first component of~\eqref{eq:peel} is linear in $h$, but the second component is clearly linear in $h$ as well. Thus, thanks to the outer-product structure of AD gradients, we can compute low rank approximation in time linear in the layer width $h$.\footnote{To declutter notation we have dropped the layer index on~$h$, as each $h \gg N$.}

We have additionally observed, that during training the true rank of
$\grad{\Loss}{\iWeight{i}}$ fluctuates and although always below $N$ may take
significantly lower values than the desired $r$ we pick for the
structured power iterations.
To skip computing noisy columns for our $\mathbf{Q}$ and $\mathbf{G}$
matrices, we stop the process when $\lVert\bfvec{g}^j -
\bfvec{g}^{j+1}\rVert_2/\lVert\bfvec{g}^{j}\rVert_2 < \theta$, where we set the threshold $\theta$ to $10^{-3}$.


\subsection{Back Propagation Through Time}

So far, we have treated rank-dAD and  SPI as applied to matrices where one dimension corresponds to the batch-size, and the other corresponds to the number of neurons in the hidden layer. In applications to recurrent networks and backpropogation through time (BPTT), we require some further consideration to clarify how we can apply rank-dAD.

In standard rank-dAD, we approximate the exact gradient by locally reducing $N \times h_i$ matrices to $r \times h_i$ matrices. These matrices are then stacked along the rank dimension, and further reduced before broadcasting back to local sites. For BPTT, we can assume that parameters for a given recurrent network are tied over the length of the sequence, and the gradient for these weights can be computed by aggregation over this sequence. Thus, a sensible extension to rank-dAD accomodates BPTT by stacking matrices into a joint batch and time dimension of size $N \cdot T$ where $T$ is the length of the sequence. We then reduce this $(N \cdot T) \times h_i$ matrix as before.

This approach to BPTT with rank-dAD allows us to obtain $O(k\times h_i)$ communication, with only a factor of $T$ increase in initial local complexity. We thus avoid expensive gradient reduction over the length of the sequence, replacing that operation completely with structured power iterations.

\section{Results}
\label{sec:results}

This section presents the results of several experiments which illustrate the communication and computational benefits of rank-dAD. We begin with small-scale experiments using a feed-forward architecture for digit classification of the MNIST data set, and then show a similar small example with a GRU-based recurrent architecture used for multivariate time-series classification on several datasets from the UEA repository. Finally, we show how rank-dAD can be used for tasks using modern transformer architectures for both sentiment analysis of the IMBD dataset, and a larger scale experiment using Vision Transformers \cite{dosovitskiy2020image} for image recognition on Cifar-10. Table \ref{tab:my_label} provides information on the three architectures which were used for experiments in this paper. 

\begin{table*}[ht]
    \centering
    \begin{tabular}{ccccp{0.3\linewidth}}
        \toprule
         \textbf{Architecture} &  \textbf{Hidden Layer Sizes} & \textbf{Sequence Length} & \textbf{Depth} & \textbf{Datasets Tested} \\\midrule
         Feed Forward & 1024,1024 & - & 2 & MNIST\\[0.5ex]
         GRU-RNN & 512,256 & 256 & 2 & Spoken Arabic Digits, PENS-SF, NATOPS, Pen-Digits\\[0.5ex]
         Vision Transformer & 128 & 50 (patch size 32) & 12 & CIFAR-10\\\bottomrule
    \end{tabular}
    \caption{Data sets and architectures tested as part of the experiments for this paper.}
    \label{tab:my_label}
\end{table*}

\begin{table}[t]
  \hfill
    \begin{tabular}{lccc}\toprule
         \textbf{Manufacturer} & \textbf{Cores} & \textbf{Memory} & \textbf{GPUs} \\\midrule
         AMD &  64 & 512 GB & 1$\times$Nvidia 2080\\
         Nvidia DGX-1  & 40 & 512 GB  & 8$\times$Nvidia V100\\
         Dell &  40 & 192 GB & 4$\times$Nvidia V100\\\bottomrule
    \end{tabular}
    \caption{Specs for the SLURM cluster used to run the experiments described in \S\ref{sec:results}.}
    \label{tab:slurm_specs}
\end{table}
All experiments were run on a SLURM cluster which submits jobs to one of the 26 machines on the same network. Specs for these machines are provided in Table~\ref{tab:slurm_specs}. For all experiments, all networks were implemented in PyTorch~1.7.1 with Python~3.8 using an Adam optimizer with a fixed learning rate of $10^{-4}$ and batch size of $64$ per site.  We performed $k=5$-fold cross-validation for all experiments, and plot the average results with error bars across these folds.  We use the gloo distributed backend for communication between nodes, with all distributed communication methods implemented in native PyTorch.

\begin{figure}[!ht]
\centering
\vskip 0.2in
  \includegraphics[width=\linewidth]{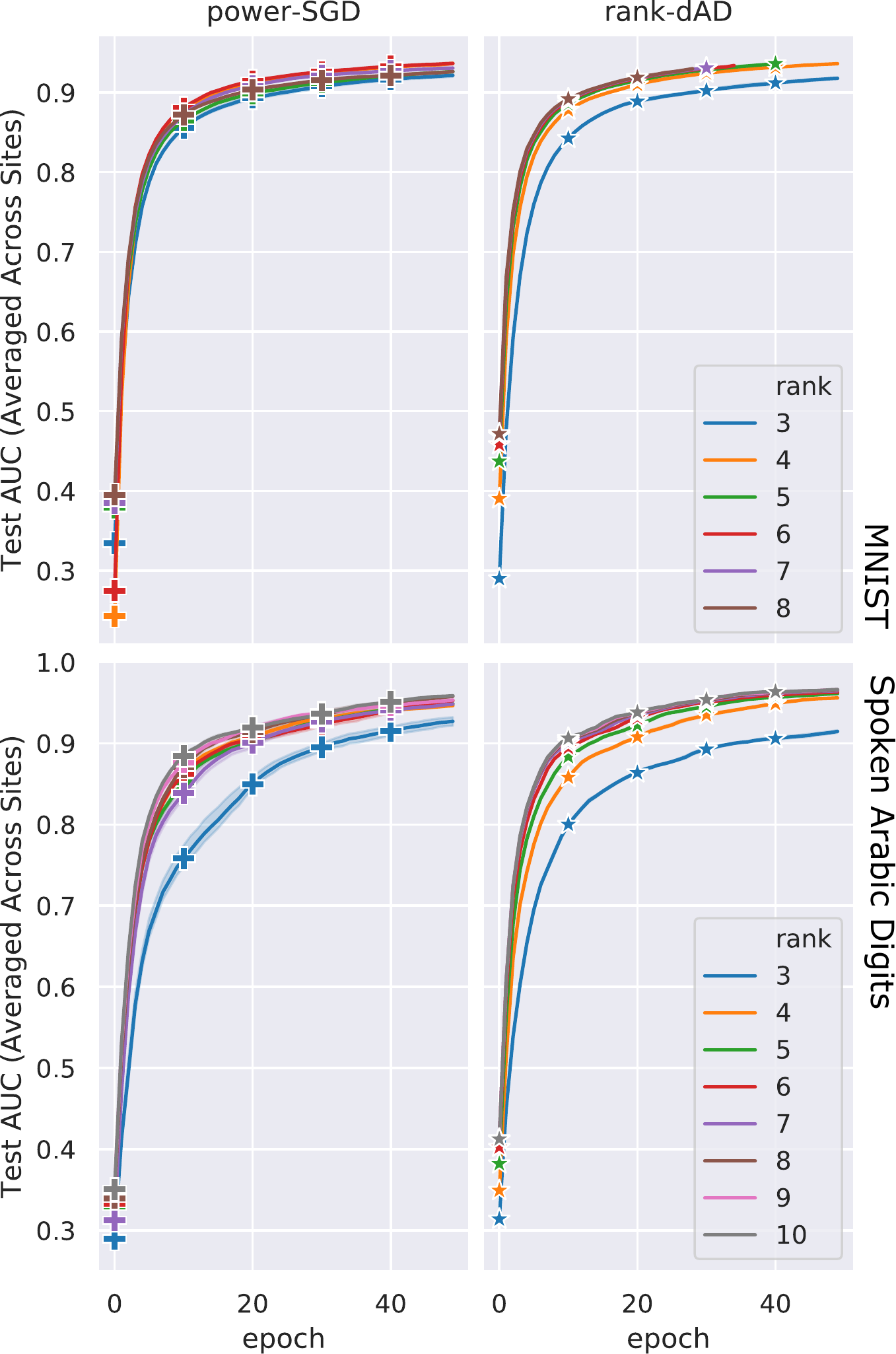}
\vspace{-10pt}
\caption{The average test AUC across sites for power-SGD and rank-dAD
  with increasing rank. Top panel: AUC over 50 epochs for
  the MNIST dataset trained on a Feed Forward Network. Bottom panel: AUC for the Spoken Arabic Digits dataset trained on a GRU-RNN.}
\label{fig:rank_auc_comparison_gru}
\vspace{-0.2in}
\end{figure}

In figure \ref{fig:rank_auc_comparison_gru}, the top two panels compare the Area-Under the Curve (AUC) for digit recognition on MNIST between power-SGD and our method. Table \ref{tab:MNIST_speedup} shows the average per-batch runtime for rank-dAD as a ratio of the average dSGD runtime. Rank-dAD provides comparable performance to dSGD, regardless of the choice of rank, and provides a speedup of between 15 and 25 times over that of vanilla dSGD. 

\begin{table*}[t]
    \small
    \centering
    \begin{tabular}{lllllllll}
    \toprule
        Mode/Sites & 4 & 6 & 8 & 10 & 12 & 14 & 16 & 18 \\\midrule
        rank-dAD & 0.063 & 0.055 & 0.049 & 0.046 & 0.039 & 0.040 & 0.037 & 0.038 \\\bottomrule
    \end{tabular}
    \caption{For a feed-forward network trained on the MNIST data set, average per-batch runtime of rank-dAD as a ratio of the per-batch runtime of dSGD. Even with only 4 sites, rank-dAD sees an over 15 times runtime decrease, and with 18 sites, over 25 times.}
    \label{tab:MNIST_speedup}
\end{table*}

\begin{figure}[ht]
  \includegraphics[trim=12 14 12 12,width=\linewidth]{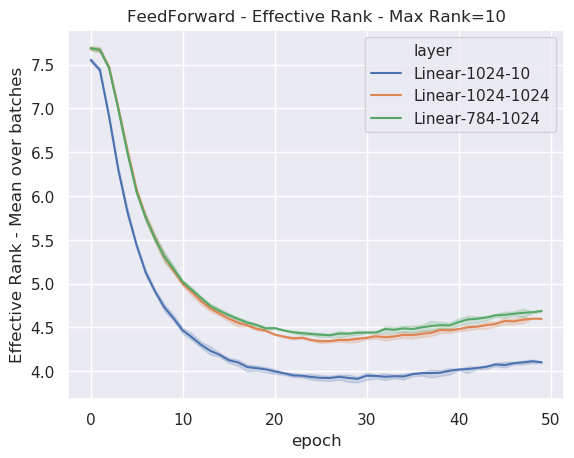}
  \vspace{-10pt}
  \caption{Effective rank for the MNIST dataset trained with a Feed Forward network across training time, where the initial maximum rank was set to 10.}
  \label{fig:mnist_effective_rank}
  \vspace{-10pt}
\end{figure}

Figure \ref{fig:mnist_effective_rank} plots the effective rank of the gradient as computed by rank-dAD during training. We notice that as the model trains, the rank needed for reliable estimation, and thus overall communication, decreases in all layers of the model. 

\begin{figure*}[ht]
  \centering
  \includegraphics[width=\linewidth]{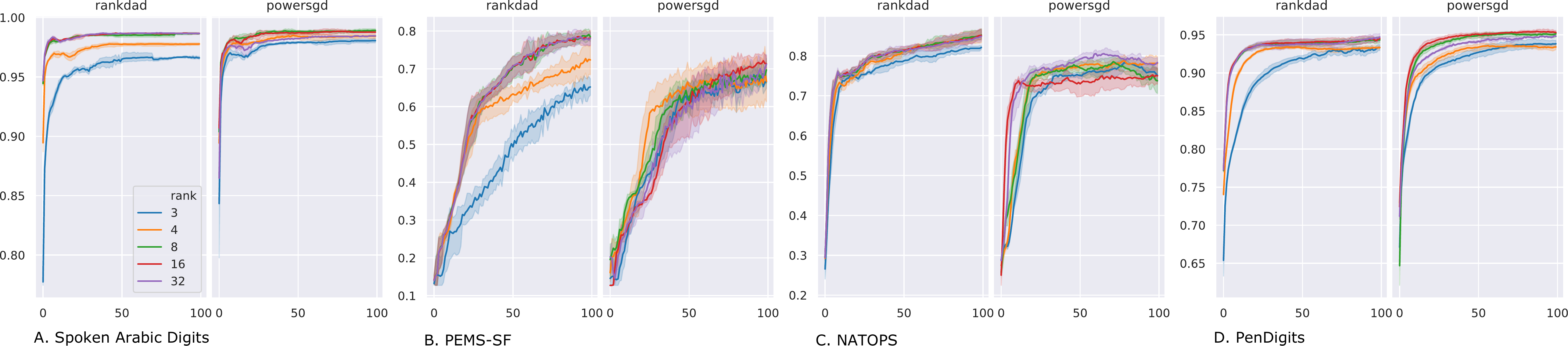}
  \vspace{-15pt}
\caption{The test AUC for a GRU-RNN trained with rank-dAD compared with the same architecture trained with PowerSGD. Each curve in the two plots provides the AUC over training for a different maximum rank. }
    \label{fig:auc_gru}
\end{figure*}

The bottom two panels of figure \ref{fig:rank_auc_comparison_gru} compare the performance of rank-dAD with power-SGD for a GRU-RNN used to classify the Spoken Arabic Digits data set for different choices of maximum effective rank. Again rank-dAD performs comparably to power-SGD during training, regardless of the choice of rank. In addition, in figure \ref{fig:auc_gru} we show the AUC compared between PowerSGD and rank-dAD for the GRU-RNN architecture on three additional datasets from the UEA repository \cite{bagnall2018uea}. 

\begin{figure}[ht]
    \includegraphics[width=\linewidth,trim=10 15 12 10]{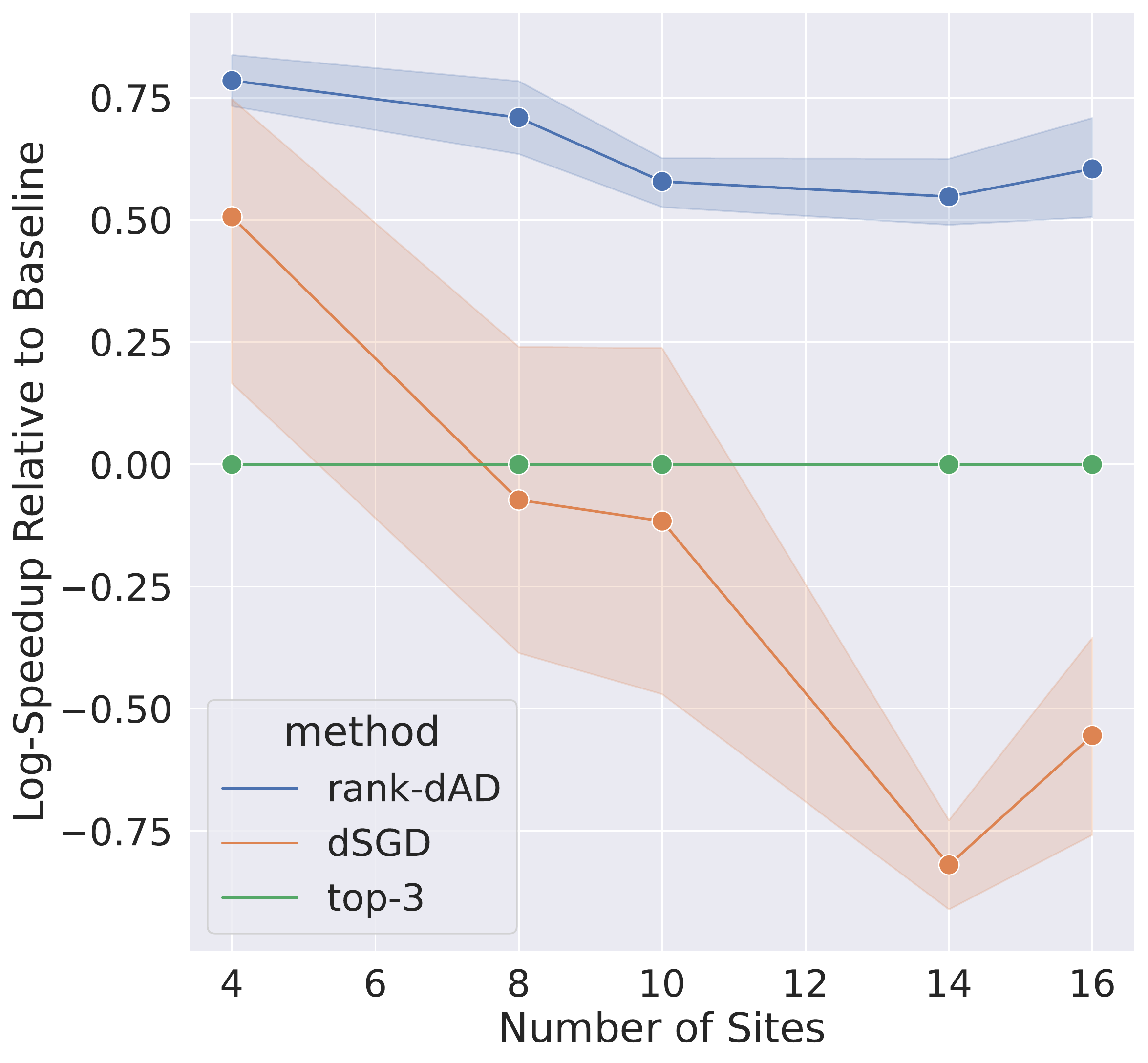}
    \vspace{-10pt}
    \caption{Using a Vision Transformer, the log speedup for achieving matching baseline AUC on CIFAR-10 trained with rank-dAD and dSGD compared to the baseline of sharing the top 3 columns from the activation and delta matrices.}
    \label{fig:vit_speedup}
    \vspace{-20pt}
\end{figure}

Finally, in figure \ref{fig:vit_speedup}, we were interested in examining how in large-scale settings with modern architectures, rank-dAD and dSGD compare with a more rudimentary baseline. As a baseline, we send only the top 3 columns from the activation and delta matrices to the aggregator and use these to compute the gradient. This amounts to effectively reducing the batch-size to 3, and discarding the majority of each batch. We illustrate in this figure how quickly it takes rank-dAD and dSGD to train to AUC comparable to this baseline. Using a Vision-Transformer as our base architecture for image recognition on CIFAR-10, our results clearly illustrate that rank-dAD provides a clear speedup over dSGD, achieving comparable AUC in a much faster runtime. Since our baseline represents a lower bound on the communication allowed between sites, we also illustrate here that rank-dAD, despite having the same communication as the baseline, is able to achieve comparble AUC faster, illustrating the clear benefit of using the rank-reduction method to leverage information from all samples.  

\section{Discussion}

This section presents an analysis of the theoretical and empirical results provided above, taking note of how each result contributes to support our claims.

\subsection{Performance}
Because dAD involves the transmission of full activations and deltas to all sites, the gradients computed by this method exactly matches those which would be computed in the pooled case, or in distributed SGD. Thus, dAD is well-suited to applications where the exact gradients are required, and its bandwidth improvements over vanilla dSGD make it the favorable choice in such cases. 

In many applications, however, low-rank approximations may be sufficient, and methods like rank-dAD may be applied. For different initial ranks, rank-dAD performs on par with PowerSGD on MNIST, and often better on the UEA datasets (see figure ~\ref{fig:rank_auc_comparison_gru}). We attribute our improved performance over PowerSGD to a stronger robustness of our low-rank approximation of the gradient in the $L_2$ sense. For the application to the transformer, rank-dAD sees a small performance hit when compared to the pooled case, which we believe is attribute to layer norms being computed locally for the distributed transformers.  

\begin{figure}[ht]
  \hfill
  \includegraphics[trim=8 8 8 8,width=0.45\textwidth]{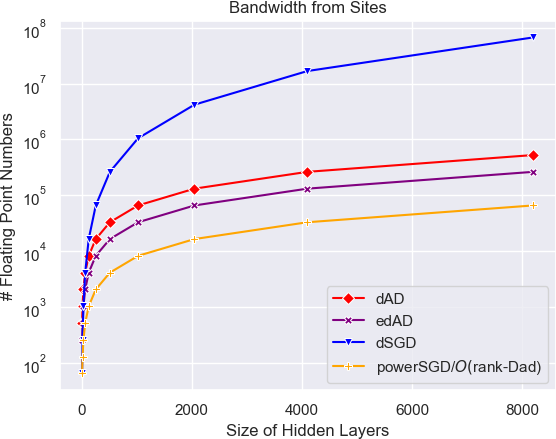}
\caption{Bandwidth per layer as a function of the layer size for the
  considered methods.}
\label{fig:bandwidth}
\vspace{-15pt}
\end{figure}





\subsection{Limitations and Future Work}

The key insight of our work is that auto-differentiation provides a
unique and useful perspective into distributed deep learning, which
can be utilized both to improve the bandwidth of distributed
algorithms and to examine the dynamics involved in
learning. Figure~\ref{fig:bandwidth} puts potential bandwidths saving
in perspective.

\paragraph{Extension to General-Purpose AD}

AD has applications to machine learning beyond deep learning, such as generic gradient-based optimization with the Hessian \citep{pearlmutter1994fast, agarwal2016second}, or Bayesian posterior inference in MCMC \citep{meyer2003stochastic}. Further work is required to look into auto-differentiation as a potentially distributable process in and of itself; however, such work would open up a much wider domain of machine learning to the insights provided here.

\paragraph{The Problem with Convolutions}
\label{sec:convolve}

In their current form, dAD and rank-dAD share input activations for the given layers in the network. For feed-forward and recurrent networks (as well as transformers), the bandwidth improvements provided over dSGD are obvious. Convolutional layers, however, present a bandwidth problem for these methods, because the size of the resulting output activations from a convolutional layer tend to be much larger in size than the number of parameters within that layer. Thus, further work is needed in examining AD applied to convolutional layers to see if bandwidth reduction is available without the addition of heuristics.




\paragraph{Effective Gradient Rank for Introspection and DNN Dynamics}
Although our initial approach was to use the unique structure provided to us by AD to compute an accurate low-rank approximation, the apparent dynamics this approach reveals beg for further empirical and theoretical analysis. The Singular Value Decomposition has been used to study the dynamics of training in networks with linear activations \citep{saxe2013exact}, and it is possible that distributed models may be provided with such an analysis for free when using our method.  By opening the black box of AD, we may get a peek into the black box of deep learning for free.

\section{Conclusions}
\label{sec:conclusion}

In this work, we took a step back from standard gradient-based methods for distributed deep learning, and presented a novel algorithm for distributed auto-differentiation (dAD). The insight that the intermediate outer product factors gathered by AD can be transferred instead of the full gradient provides a significant bandwidth reduction over full-gradient methods like dSGD without a loss in performance. Furthermore, we are able to show that the structure of AD can be further capitalized on to reduce bandwidth again by half, since for standard backpropagation, the global delta values can be back-propagated through the network as long as the activations are still shared. Finally, we push dAD even further by exploiting the explicit outer-product structure to obtain low rank approximations for the gradient in terms of low-rank versions of the intermediate statistics. With this, rank-dAD provides an intriguing method for adaptively reducing bandwidth where the chosen rank is an \textit{upper}-limit on communication. We are also able to analyze how the effective rank of the gradient changes during training, and thus obtain introspective information about the learning dynamics. It has been our goal to illustrate that auto-differentiation provides a rich landscape for further exploration into distributed deep learning. The reduction in bandwidth, intuitive algorithms, and competitive performance we have demonstrated here represent the first of potentially many benefits available to distributed machine learning practice and theory.

\section*{Acknowledgements}
Removed for anonymity

\bibliography{edad}
\bibliographystyle{icml2022}

\newpage
\appendix
\onecolumn


\end{document}